# Adaptive Affinity Propagation Clustering

Kaijun Wang[1], Junying Zhang[1], Dan Li[1], Xinna Zhang[2] and Tao Guo[1]
[1] School of computer science and engineering, Xidian University, Xi'an 710071, P. R. China.
[2] China Jiliang College, Hangzhou 310018, P. R. China
Email: sunice9@yahoo.com

**Abstract**  Affinity propagation clustering (AP) has two limitations: it is hard to know what value of parameter 'preference' can yield an optimal clustering solution, and oscillations cannot be eliminated automatically if occur. The adaptive AP method is proposed to overcome these limitations, including adaptive scanning of preferences to search space of the number of clusters for finding the optimal clustering solution, adaptive adjustment of damping factors to eliminate oscillations, and adaptive escaping from oscillations when the damping adjustment technique fails. Experimental results on simulated and real data sets show that the adaptive AP is effective and can outperform AP in quality of clustering results.

## 1  Introduction

Affinity propagation clustering (AP) [1] is a fast clustering algorithm especially in the case of large number of clusters [2], and has some advantages: speed, general applicability and good performance. AP works based on similarities between pairs of data points (or $n \times n$ similarity matrix $S$ for $n$ data points), and simultaneously considers all the data points as potential cluster centers (called exemplars). To find appropriate exemplars, AP accumulates evidence "responsibility" $R(i,k)$ from data point $i$ for how well-suited point $k$ is to serve as the exemplar for point $i$, and accumulates evidence "availability" $A(i,k)$ from candidate exemplar point $k$ for how appropriate it would be for point $i$ to choose point $k$ as its exemplar. From the view of evidence, larger the $R(:,k)+A(:,k)$, more probability the point $k$ as a final cluster center. Based on evidence accumulation, AP searches for clusters through an iterative process until a high-quality set of exemplars and corresponding clusters emerges. In the iterative process, identified exemplars start from the maximum $n$ exemplars to fewer exemplars until $m$ exemplars appear and are unchanging any more (or AP algorithm converges). The $m$ clusters found based on $m$ exemplars are the clustering solution of AP.

There are two important parameters in AP: the preferences ($p$) in diagonal of similarity matrix $S$ and the damping factor (*lam*). The preference parameter $p(i)$ (its initial value is negative) indicates the preference that data point $i$ be chosen as a cluster center, and influences output clusters and the number of clusters (NC). The main formula in the AP algorithm are: $R(i,k) = S(i,k) - \max\{A(i,j)+S(i,j)\}$, where $j \in \{1,2,\ldots,n\}$ but $j \neq k$; and $A(i,k) = \min\{0, R(k,k)+\text{sum}\{\max(0,R(j,k))\}\}$, where $j \in \{1,2,\ldots,n\}$ but $j \neq i$ and $j \neq k$; and $p$ appears in $R(k,k)=p(k)-\max\{A(k,j)+S(k,j)\}$. Hence, the $p$ influences which and how many exemplars will win as final cluster centers, i.e., when $p(k)$ is larger, $R(k,k)$ and $A(i,k)$ are larger, so that it has more probability that the point $k$ is as a final cluster center. This means that the number of identified clusters is increased or decreased by adjusting $p$ correspondingly, and usually a good choice is to set all the $p(i)$ to be the median (*pm*) of all the similarities between data points [1]. However, the *pm* can not lead to an optimal clustering solution in many cases, since the *pm* is not given on the basis of the optimal cluster structure of a data set. Furthermore, there is no exact corresponding relation between the $p$ and output NC. Therefore, how to find an optimal clustering solution is an unsolved problem for using AP algorithm.

In each iterative step $i$, $R$ and $A$ are updated with the one in last iteration, i.e., $R_i = (1-lam) \times R_i + lam \times R_{i-1}$, $A_i = (1-lam) \times A_i + lam \times A_{i-1}$, where damping factor $lam \in [0,1]$ and default $lam=0.5$. Another function of the damping factor is to improve convergence when AP fails to converge on account of oscillations (or identified exemplars are in periodic variation), where *lam* needs to be increased to eliminate oscillations [1]. In un-convergent cases, we have to increase *lam* manually and gradually and rerun AP until the algorithm converges. Another choice is to use a big damping factor close to 1 to eliminate oscillations, but AP will run very slow. Both choices may consume plenty of time, especially for a large data set. Hence, it is an important problem for AP: how to automatically eliminate oscillations when oscillations occur?

To solve these problems, we propose adaptive AP, including: adaptive adjustment of the damping factor to eliminate oscillations (called adaptive damping), adaptive escaping oscillations by decreasing $p$ when adaptive damping method fails (called adaptive escape), and adaptive searching the space of $p$ to find out the optimal clustering solution suitable to a data set (called adaptive preference scanning). The adaptive AP is proposed in Section 2, and experimental results are in Section 3. Finally, Section 4 gives the conclusion.

The Matlab codes of the adaptive AP are available from ref. [10].

## 2  Adaptive Affinity Propagation

In this section, the adaptive damping and escape methods are discussed first to eliminate oscillations, and then the adaptive scanning of *p* is designed. Finally, a cluster validity method is adopted to find the optimal clustering solution. It is noted that the same initial value is assigned to all the *p*(*i*) in the diagonal of matrix *S*.

When oscillations occur and AP fails to converge, our target is to both eliminate oscillations and keep the speed of the algorithm. Although the *lam* near 1 has more probability for oscillation elimination, the *R* and *A* are updated very slow and much more time is needed to run AP. It is a better choice to check the effect of oscillation elimination while increasing *lam* gradually. Following this idea, the adaptive damping method is designed: (1) detect whether oscillation occurs in an iteration of the AP algorithm; (2) the *lam* is increased once by a step such as 0.05 if oscillations are detected, otherwise go to (1); (3) the iteration continues *w* times; (4) repeat these steps until the algorithm meets preset stop condition.

It is a key to detect any oscillation while the algorithm runs, but features of oscillations are too complex to be described. Then, we turn to describe/define non-oscillation features: the number of identified exemplars is decreasing or unchanging during the iterative process. This definition is reasonable, since the decreasing and unchanging are the features that the algorithm is going to convergence. A moving monitoring window *Kb*(*i*) (window size *w*) is used to record whether non-oscillation features appear in a sequence of iterations, e.g., *Kb*(*i*)=1 when non-oscillation features appear in iteration *i*, otherwise *Kb*(*i*)=0. The criterion of whether oscillations occur is designed as follows: oscillations occur if the number of non-oscillation features in the monitoring window is less than two thirds of window size. This is a tolerant design that considers tolerating occasionally random vibrations in a short time and vibrations in initial iterations. Thus, the above monitoring -adjusting technique realizes adaptive adjustment of *lam* and leads AP to convergence.

If it fails to depress oscillations by increasing *lam* (e.g., *lam* is increased to 0.85 or higher), an adaptive escape technique will be designed to avoid oscillations. That large *lam* brings little effect suggests that oscillations are pertinacious under the given *p*, so the alternative is to decrease *p* away from the given *p* to escape from oscillations. This escape method is workable due to that it works together with adaptive scanning of *p* discussed below, different from AP that works under a fixed *p*.

The adaptive escape technique is designed as follows: when oscillations occur and *lam*≥0.85 in the iterative process, decreasing *p* gradually until oscillations disappear. This technique is added in the step (2) of the adaptive damping method: if oscillations occur, increasing *lam* by a step such as 0.05; if *lam*≥0.85, decreasing *p* by step *ps*, otherwise go to step (1) of the adaptive damping method. Both adaptive damping and adaptive escape techniques are used to eliminate oscillations at the same time. The monitoring window size *w*=40 is appropriate as per our experiences (occasionally random vibrations and tolerant vibrations in initial iterations will be caught under too small *w* and AP runs slowly under too big *w*). The pseudo codes of adaptive damping and adaptive escape are listed in Table 1 (where *maxits* and *ps* will be set in Table 2).

**Table 1**. Procedure of adaptive damping and adaptive escape

```
Initialization: damping factor lam←0.5, monitoring window size w←40,
parameter w₂←w/8, max iteration times maxits, decreasing step ps.
for i←1 to maxits do
    Kset(i) ← K          △ K is the number of exemplars
    Km(i)←mean(Kset(i-w₂:i))
    if Km(i)-Km(i-1) < 0  then Kd←1   △ record the decrease of K
    Kc←∑ⱼ |Kset(i)-Kset(j)|   △ record the unchanged of K, j∈i-w₂ to i-1
    △ record the decrease and unchangeableness of K in monitoring
        window Kb:
    if Kd = 1 or Kc = 0
        then   Kb(j)←1      △ j is the remainder of i/w
        else   Kb(j)←0      △ j is the remainder of i/w
    Ks←∑ⱼ Kb(j)            △ j∈1 to w
    if Ks < 2w/3
        then   lam←lam+0.05
            if lam >= 0.85   then   p←p+ps
```

The number of identified clusters depends on input *p*, but it is unknown which value of *p* will give best clustering solution for a given data set. Generally, cluster validation techniques (usually based on validation indices) [3] are used to evaluate which clustering solution is optimal for a data set. AP algorithm need give a series of clustering solutions with different NCs, among which the optimal clustering solution is found by a cluster validation index. There is no exact corresponding relation between the *p* and output NC, so we design the method of scanning space of *p* to obtain different NCs.

The adaptive *p*-scanning technique is designed as follows: (1) specify a large *p* to start the algorithm; (2) an iteration runs and gives *K* exemplars; (3) check whether *K* exemplars converge (the condition is that every exemplar satisfies preset continuously unchanging times *v*); (4) go to step (5) if *K* exemplars converge, otherwise go to step (2); (5) decrease the *p* by step *ps* if *K* exemplars converge too in additional *dy* iterations (this is for more reliable convergence), otherwise go to step (2); (6) go to step (2).

Thus, a series of clustering results with different NCs can be gained through scanning *p*, and the scanning of *p* space is designed inside the iterative process to keep the advantage of speed. To avoid possible repeated computation, in the *p*-scanning process we continue to calculate *R(i,k)* and *A(i,k)* based on (or using) the current values of *R(i,j)* and *A(i,j)* after each reduction of *p* (then *S(i,i)=p(i)* is changed but other elements of *S* are unchanged).

It is the key to select a proper decreasing step for the adaptive *p*-scanning technology. According to our experience, the decreasing step may be set to be *ps=pm/100*. This is a compromise design, which considers both cases: the algorithm runs slow when |*ps*| is too small, and the algorithm possibly miss the NC of the inherent cluster structure when |*ps*| is too big. Nevertheless, this fixed decreasing step cannot meet the different cases of big NC and small NC. Considering that big NC is more sensitive to *ps* than that of small NC, we design the adaptive decreasing step: *ps=0.01pm/q*, where decreasing parameter $q = 0.1\sqrt{K+50}$. Thus, the *q* is adjusted dynamically with *K*, and the *ps* is smaller when *K* is bigger, while the *ps* is larger when *K* is smaller.

In order to check whether the convergence condition is satisfied, another monitoring window *B* (similar to that in adaptive damping method) is adopted to record the continuously unchanging times *v* of *K* exemplar, and the window size is set to be *v*=40, which is consistent with default convergence times 50 in AP [1] (*v*=40 pluses delay times of 10).

It is important to specify the scanning scope, and a smaller scope is preferred for less running time. The *p* space [-∞, 0] corresponds to NC space [1, *n*]. For the clustering of *n* data points, it is reasonable to regard the square of *n* as the upper limit of the optimal NC [4]. In the following experiments we find: $K_1$ at the first convergence equals or is over $\sqrt{n}$ when initial *p=pm/2* is set, and the NCs searched by the algorithm are much bigger than $\sqrt{n}$ (since every data point is regarded as an exemplar when AP starts). Hence, we set initial *p=pm/2*. The minimal NC (*K*=2) determines the lower limit of *p*, i.e., reducing *p* until *K*=2. The large *maxits*=50000 is set so that the maximal iteration times *maxits* does not influence whether the algorithm reaches *K*=2.

Finally, an acceleration technique of *p*-scanning is needed to save running time. As some NCs correspond to a large scope of *p*, the large reduction of *p* is needed to change NC. In this case, we may increase the decreasing step of *p* to obtain smaller NCs rapidly. The acceleration technique of *p*-scanning is designed as follows: (1) the iteration runs once; (2) check whether *K* exemplars converge; if yes, go to (3), otherwise set *b*=0 and go to (1); (3) the iteration continues *dy*=10 times; (4) check whether *K* exemplars converge; if yes, set *b=b+*1, otherwise go to (1); (5) set *p = p+b×ps*, and go to (3). The pseudo codes of the adaptive *p*-scanning technology are listed in the Table 2.

Now the adaptive AP gives clustering solutions with different NCs through the *p*-scanning process, and then cluster validation technique is used to evaluate quality of these solutions. It is the validity indices that are usually used to evaluate quality of clustering results and to evaluate which clustering solution is the optimal for the data set. Among many validity indices, Silhouette index, which reflects the compactness and separation of clusters, is widely-used and has good performance on NC estimation for obvious cluster structures. It is applicable to both the estimation of the optimal NC and evaluation of clustering quality. Hence, we adopt Silhouette index, as an illustration, to find the optimal clustering solution.

Let a data set with *n* samples be divided to *k* clusters $C_i$ (*i*=1~*k*), *a(t)* be average dissimilarity of sample *t* of $C_j$ to all other samples in $C_j$, *d(t,$C_i$)* be average dissimilarity of sample *t* of $C_j$ to all samples in another cluster $C_i$, then *b(t)*=min{*d(t,$C_i$)*}, *i*=1~*k*, $i \neq j$. The Silhouette formula for sample *t* is [3]:

$$Sil(t) = \frac{b(t) - a(t)}{\max\{a(t), b(t)\}} \quad (1)$$

With *Sil*(*t*) for each sample, overall average silhouette *Sil* for *n* samples of the data set is obtained directly. The largest overall average silhouette indicates the best clustering quality and the optimal NC [3]. Using formula (1), a series of *Sil* values corresponding to clustering solutions under different NCs are calculated, and the optimal clustering solution is found at the largest *Sil*.

**Table 2**. Procedure of adaptive *p*-scanning for searching NC space

```
Initialization: p←pm/2, ps←pm/100, b←0, v←40, dy←10,
            nits←0, maxits←50000.
for i←1 to maxits do
    Kset(i)←K              △ K is the number of exemplars
    if point k is the exemplar
        then   B(k,j)←1    △ j is the remainder of i/v
        else   B(k,j)←0    △ j is the remainder of i/v
    if there are K exemplars that make ∑_j B(k,j) = v
        then   Hdown←1     △ K exemplars converge
        else   Hdown←0, b←0, nits←0
    nits←nits+1
    if   Hdown = 1 and nits >= dy
        then   b←b+1
               q ← 0.1√(K+50)
               p←p+b×ps/q
               nits←0
    if K <= 2
        then stop
```

## 3  Experimental Results

This section compares the clustering performance between adaptive AP method (adAP) and AP algorithm (AP). The items of clustering performance include: whether adAP can eliminate oscillations (if oscillations occur) automatically so as to give correct clustering results, whether adAP can give correct clustering results based on the Silhouette index (or cluster validation technique). The adAP and AP use same initial *lam*=0.5 (but *lam*=0.8 in Travelroute experiment), and AP uses fixed *p*=*pm* and *maxits*=2000. For Document and Travelroute experiments, both methods use fixed *p* from prior knowledge [1].

Let a data set be an *n*×*d* matrix $X=\{x_i\}$, where $x_i$ is *d*-dimensional. For general data, the similarity between sample $x_i$ and $x_j$ is $b_{ij} = -\|x_i - x_j\|^2$ based on Euclidean distances, while for gene expression data the Pearson coefficients are used as similarity measure, i.e., the linear relationship between two samples $x_i$ and $x_j$ (their means $\bar{x}_i$ and $\bar{x}_j$ ) is:

$$R(i,j) = \frac{\sum_{l=1}^{d}(x_{il}-\bar{x}_i)(x_{jl}-\bar{x}_j)}{\sqrt{\sum_{l=1}^{d}(x_{il}-\bar{x}_i)^2}\sqrt{\sum_{l=1}^{d}(x_{jl}-\bar{x}_j)^2}} \qquad (2)$$

To avoid possible calculation confusion from negative values, the $R(i,j) \in [-1,1]$ is transformed to $R(i,j) = 1-(1+R(i,j))/2$. Thus, Pearson coefficients are transformed to positive Pearson distance $R(i,j) \in [0,1]$ (bigger the value, farther the two samples), and the similarity between sample $x_i$ and $x_j$ is $b_{ij} = -R(i,j)$.

Twelve data sets in Table 3 are used in the experiments, where the first eight data sets have known class labels. Their features include: far and close well-separated clusters, slight overlapping clusters, tight clusters and loose clusters. The first four data sets are simulated data, while other data sets are real data. The Yeast and NCI60 are gene expression data, and a subset of dataset Exons is used, i.e., the first 3499 samples and the last one (= 3500 samples) from 75067 samples are used.

**Table 3**. Features of data sets

| Data sets | Features of cluster structures | Number of clusters | Number of samples | Dimensions | Source |
|---|---|---|---|---|---|
| 3k2lap | overlap, loose | 3 | 300 | 2 | [5] |
| 5k8close | close, loose | 5 | 1000 | 8 | [6] |
| 14k10close | close, loose | 14 | 480 | 10 | [5] |
| 22k10far | far, tight | 22 | 790 | 10 | [5] |
| Ionosphere | overlap, loose | 2 | 351 | 4 | [7] |
| Wine | overlap, loose | 3 | 178 | 3 | [7] |
| Yeast | far, loose | 4 | 208 | 79 | [8] |
| NCI60 | overlap, loose | 8 | 58 | 20 | [9] |
| FaceImage | overlap | 100 | 900 | 50×50 | [1] |
| Document | / | 4 | 125 | / | [1] |
| Travelroute | / | 7 | 456 | 3 | [1] |
| Exons | / | / | 3500 | 12 | [1] |

The clustering results of adAP and AP are listed in Table 4, where 'adAP error' denotes error rates of adAP solutions (compared with true class labels), 'yes' in 'adAP eliminate oscillations' denotes that oscillations occur and adAP eliminates them automatically, 'adAP time' and 'AP time' denote the running time of Matlab programs of adAP and AP respectively in a same computer (Intel CPU 3.60GHz, 2GB), and FM denotes Fowlkes-Mallows index [3]. FM with values in [0,1] measures agreement between a clustering solution and true class labels, and bigger the value better the agreement, e.g., FM value is 1 when the solution and true labels are the same. FM index is used to evaluate clustering quality when the NC of a clustering solution is different from true NC. In addition, the last four datasets have no true class labels so that there is no error rate and FM value; for Exons, the values in FM denote correct rates of found exons.

**Table 4**. Clustering results of adAP and AP

| Data sets | known NC | adAP NC | adAP error (%) | adAP FM | adAP time (s) | adAP eliminate oscillations | AP NC | AP FM | AP time (s) |
|---|---|---|---|---|---|---|---|---|---|
| 3k2lap | 3 | 3 | 7.67 | 0.85 | 144.0 | / | 16 | 0.39 | 2.1 |
| 5k8close | 5 | 5 | 0 | 1.00 | 1851.0 | / | 17 | 0.56 | 31.2 |
| 14k10close | 14 | 14 | 0 | 1.00 | 275.5 | / | 15 | 0.97 | 6.0 |
| 22k10far | 22 | 22 | 0 | 1.00 | 1125.9 | yes | 168 | 0.80 | 307.8 |
| Ionosphere | 2 | 2 | 17.4 | 0.75 | 445.3 | yes | 28 | 0.43 | 56.8 |
| Wine | 3 | 3 | 10.7 | 0.80 | 34.5 | / | 11 | 0.46 | 0.5 |
| Yeast | 4 | 4 | 3.37 | 0.97 | 54.7 | / | 11 | 0.66 | 0.9 |
| NCI60 | 8 | 8 | / | 0.56 | 12.9 | / | 9 | 0.48 | 0.1 |
| FaceImage | 100 | 102 | / | / | 3701.2 | / | 103 | / | 14.5 |
| Document | 4 | 4 | / | / | 0.3 | / | 4 | / | 0.2 |
| Travelroute | 7 | 7 | / | / | 24.7 | / | 7 | / | 20.0 |
| Exons | / | 102 | / | 32.8% | 83073.7 | / | 37 | 22.4% | 996.0 |

In Table 4 one can see: for all the datasets except the last four datasets, adAP gives correct NC in all the cases, while AP fails in all the cases; FM values of adAP are higher than that of AP, indicating that adAP gives better clustering quality than AP; and the oscillations lead AP to poor solutions for 22k10far and Ionosphere. The clustering task is to find representative sentences (or cluster centers) for Document data, and both adAP and AP find the same four representative sentences; and the task is to find the appropriate airport (or cluster centers) as airport hub for Travelroute data, and both adAP and AP find the same seven airports. The clustering task is to find the cluster of exons for Exons data, which is realized by finding the cluster of non-exon (it is known that the last sample is non-exon), and adAP has higher identification rate of exons than AP. Some clusters of 900 face images from 100 persons are overlapping on account of semblable faces for FaceImage data, so that the separability of the 100 clusters is not good; in this case, adAP yields a better result of 102 clusters than AP. The experimental results show that adaptive AP gives correct clustering results based on Silhouette index, and eliminates oscillations automatically. These results demonstrate that the adaptive damping, adaptive escape and adaptive preference scanning techniques in adaptive AP method are effective, resulting in better performance of adAP than original AP.

# 4  Conclusion

The proposed adaptive AP uses adaptive preference scanning to search space of the number of clusters, and finds the optimal clustering solution suitable to a data set by the cluster validation technique. Moreover, in adaptive AP the adaptive damping is designed to eliminate oscillations automatically instead of manually, and the adaptive escaping is developed to eliminate oscillations when the damping technique fails. With these adaptive techniques, adaptive AP can outperform or equal AP algorithm in clustering quality and oscillation elimination. In addition, it is worth further research that which validity method combined with adaptive AP is more appropriate for a data set with complex cluster structures such as overlapping clusters.

# References

1. Frey B J, Dueck D. Clustering by Passing Messages between Data Points. *Science*, 2007, 315(5814), 972-976. http://www.psi.toronto.edu/affinitypropagation
2. Karen K. Affinity program slashes computing times. [online] available: http://www.news.utoronto.ca/bin6/070215-2952.asp, Oct. 25, 2007.
3. Dudoit S, Fridlyand J. A prediction-based resampling method for estimating the number of clusters in a dataset. *Genome Biology*, 2002, 3(7): 0036.1-0036.21.
4. Yu J, Cheng Q S. The upper bound of the optimal number of clusters in fuzzy clustering. *Science in china*, Ser. F, 2001, 44(2): 119~125.
5. Dembélé D, Kastner P. Fuzzy C-means method for clustering microarray data, *Bioinformatics*, 2003, 19(8): 973-980.
6. Strehl, A. Relationship-based Clustering and Cluster Ensembles for High-dimensional Data Mining. Ph.D thesis, The University of Texas at Austin, May 2002.
7. Blake C L, Merz C J. UCI repository of machine learning databases, [online] available: http://mlearn.ics.uci.edu/MLRepository.html
8. Ben-Hur A, Guyon I A. Stability based method for discovering structure in clustered data. In: Proceedings of the 7th Pacific Symposium on Biocomputing. Lihue, Hawaii, USA, 2002, 6-17.
9. Ross D T, Scherf U, Eisen M B, et al. Systematic variation in gene expression patterns in human cancer cell lines. *Nature Genetics*, 2000, 24(3): 227-234.
10. http://www.mathworks.com/matlabcentral/fileexchange/loadAuthor.do?objectType=author&objectId=1095267